\pdfoutput=1

\documentclass[11pt]{article}

\usepackage[final]{acl}

\usepackage{times}
\usepackage{latexsym}

\usepackage[T1]{fontenc}
\usepackage{tabularx,multirow,makecell}
\usepackage[utf8]{inputenc}

\usepackage{microtype}

\usepackage{multirow}
\usepackage{inconsolata}

\usepackage{graphicx}
\usepackage{amsmath}
\usepackage{booktabs}
\usepackage{array}
\usepackage{enumitem}
\usepackage{inconsolata}
\usepackage{marvosym}
\usepackage{graphicx}
\usepackage{stfloats}
\usepackage{arydshln}
\usepackage{float}  
\usepackage{cuted}
\usepackage{subcaption}
\usepackage{siunitx}
\newcolumntype{Y}{>{\raggedright\arraybackslash}p{0.22\linewidth}}
\usepackage[table]{xcolor}
\usepackage{tabularx}  
\usepackage{tocloft}
\usepackage{etoc}
\usepackage{hyperref}
\usepackage{xurl} 
\usepackage{titletoc}
\usepackage{todonotes}
\usepackage{makecell}
\usepackage{tablefootnote}
\usepackage[skip=6pt]{caption}
\usepackage{placeins} 
\usepackage{tcolorbox}
\tcbuselibrary{skins}
\usepackage{subcaption}
\definecolor{claimbg}{HTML}{FFF2CC}%

\usepackage{bm}
\definecolor{darkgreen}{RGB}{0,100,0}
\definecolor{mygreen}{RGB}{0,120,0} 
\definecolor{finetunegray}{gray}{0.93}

\definecolor{Orange}{HTML}{E69F00} 
\definecolor{Yellow}{HTML}{F0E442}  
\definecolor{BluishGreen}{HTML}{009E73}

\definecolor{score2bg}{HTML}{FFF7E6}  
\definecolor{score1bg}{HTML}{E8F5FF}  
\definecolor{score0bg}{HTML}{E8FFE8}  

\usepackage{tcolorbox}
\tcbset{
  myTableBox/.style={
    colback=#1!20,        
    colframe=#1!60,       
    arc=1mm,              
    boxrule=0.4pt,        
    left=1mm,             
    right=1mm,            
    top=1mm,              
    bottom=1mm,           
  }
}

\usepackage[table]{xcolor}

\title{RIGOURATE: Quantifying Scientific Exaggeration with Evidence-Aligned Claim Evaluation}

\author{Joseph James$^1$,~~~ Chenghao Xiao$^2$, ~~~Yucheng Li$^3$ \\ 
\textbf{Nafise Sadat Moosavi$^{1}$, ~~ Chenghua Lin$^{4}$\thanks{\,Corresponding author}}\\
     $^1$The University of Sheffield, UK~~ 
    $^2$ Durham University, UK \\
    $^3$ University of Surrey, UK~~~ 
    $^4$ The University of Manchester, UK\\
    \texttt{jhfjames1@sheffield.ac.uk} ~~~
    \texttt{chenghua.lin@manchester.ac.uk}}

\begin{document}
\maketitle

\begin{abstract}
Scientific rigour tends to be sidelined in favour of bold statements, leading authors to overstate claims beyond what their results support.  We present RIGOURATE, a two-stage multimodal framework that retrieves supporting evidence from a paper’s body and assigns each claim an overstatement score. The framework consists of a dataset of over 10K claim–evidence sets from ICLR and NeurIPS papers, annotated using eight LLMs, with overstatement scores calibrated using peer-review comments and validated through human evaluation. It employes a fine-tuned reranker for evidence retrieval and a fine-tuned model to predict overstatement scores with justification. Compared to strong baselines, RIGOURATE enables improved evidence retrieval and overstatement detection. Overall, our work operationalises evidential proportionality and supports clearer, more transparent scientific communication.
All code, models, and annotation scripts will be made publicly available [Github/HF Link].
\end{abstract}

\section{Introduction}

Effective scientific writing and reviewing demands not only the clear presentation of novel ideas but also the rigorous grounding of findings that support them. In many research papers, authors are incentivised to write abstracts and introductions that capture readers’ attention by showcasing their contributions in an eye‑catching manner \citep{ bavdekar2015writing, rahman2017rhetorical, kawase2018rhetorical,hyland2021our, intemann2022understanding,stavrova2025scientific}. However, when such claims are exaggerated they can mislead the reader if robust evidence is not provided in later sections. In a rapidly evolving field such as Machine Learning \citep{pineau2021improving}, this dynamic has fostered a ``Publish or Perish'' environment \citep{de2005publish, rawat2014publish}. The pressure to publish quickly can lead authors to prioritise speed over scientific rigour, thus accumulating ``scientific debt'' \citep{nityasya-etal-2023-scientific}, where well‑grounded, reproducible work is sidelined in favour of rapid publication, diminishing the impact of scientific research.

We focus on the phenomenon of overstatement: rhetorical exaggeration in which the wording of a claim amplifies its strength beyond what the paper’s evidence supports. Our work targets a distinct aspect of fact-checking, namely quantifying degrees of overstatement, where claims are rarely outright false but often exceed what the available evidence warrants. Rather than making binary true/false judgments, we assess whether the strength and scope of a claim are proportionally grounded in the paper’s own methods and results. 
A claim is overstated when it presents an inflated representation that is not adequately supported by the paper, 
for example due to limited evidence or unjustified generalisation.
This perspective highlights how linguistic choices and rhetorical emphasis can shape readers’ perceptions of contribution even before they encounter the technical sections, thereby influencing how the communicative rigour of a claim and its alignment with the supporting evidence are perceived. Building on this motivation, we define the task of intra-paper overstatement detection, which evaluates whether claims in the abstract and introduction are proportionally supported by evidence presented in the remainder of the paper.

To address this problem, we introduce RIGOURATE, a multimodal, review‑informed, automated framework that tackles two tasks: (i) evidence retrieval and (ii) overstatement detection with reasoning. We collect papers and reviews from ICLR and NeurIPS hosted on OpenReview,
a venue with openly accessible reviews covering diverse NLP and Machine‑Learning topics. We employ a panel of large language model (LLM) annotators to identify claims that are the authors’ own statements. Then  we extracted potential evidence spanning text, figures, and tables, classifying each passage as either \emph{relevant} (directly related to the claim) or \emph{irrelevant}. Since the core issue is the degree of support rather than direct refutation, we focus on how thoroughly the available evidence grounds each claim.
We assign every claim a continuos overstatement score ranging from 0 to 1. 
The LLM annotators first generates a score for each claim–evidence set, and then we incorporate each review comment as additional context. This design reduces sensitivity to individual LLM and reviewer perspectives, with annotation quality validated through targeted human evaluation.

We use RIGOURATE to assess the feasibility and validity of intra-paper overstatement detection. The evaluation consists of two components: (i) evidence retrieval, which tests whether models can reliably identify passages, figures, and tables that support a given claim, and (ii) overstatement detection, which assesses whether models can estimate the degree to which a claim is proportionally supported by that evidence. We adopt a range of existing state-of-the-art reranker and multimodal models to this setting. The results show that fine-tuning enables these models to meaningfully learn the task signal, yielding consistent improvements across retrieval metrics and more accurate overstatement scoring compared to base zero-shot models. These findings indicate that claim–evidence alignment within a paper is a learnable and evaluable problem, and that RIGOURATE provides a practical framework for computationally assessing rhetorical overstatement via evidential proportionality, supporting automated, evidence-based evaluation of research claims and more transparent scientific communication. To summarise, our main contributions are three-fold:
\begin{itemize}[noitemsep,topsep=1pt,parsep=1pt,partopsep=1pt]
\item A review‑informed framework that automatically extracts and scores multimodal claim–evidence sets within research papers.
\item Showing that intra-paper claim–evidence alignment is a learnable task, with fine-tuning consistently improving performance on evidence retrieval and overstatement scoring.
\item A case study showing that overstatement often stems from missing substantive detail and surface-level phrasing.
\end{itemize}

\section{Related Work}
\noindent\textbf{2.1~ Scientific Rigour.}~ The integrity of scientific research is increasingly challenged by issues of reproducibility, alongside broader concerns about how claims are framed and interpreted. While reproducibility and reporting practices have been widely studied \citep{iclm_reproducibility_workshop, ml_reproducibility_challenge}, comparatively less attention has been paid to the alignment between the strength of scientific claims and the evidence presented to support them. The framing of a paper’s contributions can strongly shape perceptions of its value and rigour, allowing novel but non-replicable claims to gain prominence when independent verification is limited \citep{salager1994hedges, ferrari2019we, gustafson2020review,serra2021nonreplicable, james-etal-2024-rigour}. Although such non-replicability is not necessarily caused by exaggerated framing, it reduces opportunities for empirical scrutiny, allowing weakly supported claims to persist and thereby undermining research credibility \citep{raghupathi2022reproducibility}.



The NeurIPS and ARR checklist have addressed several challenges, including insufficient exploration of variables, poor documentation, and a lack of reporting of crucial details needed to replicate results \citep{pineau2021improving, arr}. While the checklist approach has improved the quality of research, it requires reviewers to manually validate whether the claims made are backed up by sufficient evidence. By detecting overstated claims automatically, our work aims to support authors in presenting their work accurately and to provide a framework for evaluating the rigour in research claims. Evidence that this remains challenging is provided by peer review analyses showing that even highly rated papers often receive requests for additional experiments, indicating unresolved gaps between claims and supporting evidence \citep{wang2023have}. This motivates the need for systematic methods to assess claim–evidence alignment.

\noindent\textbf{2.2~ Scientific Claim Verification.}~ 
The increasing number of publications requires the development of automated methods for verifying research claims. Scientific fact verification, which aims to assess the accuracy of scientific statements, often relies on \textit{external knowledge} to support or refute claims \citep{wadden2022scifact, vladika2023scientific, dmonte2024claim}. However, the use of abstracts as the primary source of evidence is a key limitation. As the abstract can also be overstated or omit detailed information, and so it is important to evaluate the evidence in the main body of the paper to determine if the statements made in the abstracts are well-supported.

Recent work has highlighted the importance of grounding claims in paper-internal evidence, with \citet{chan2024overview} collecting claims linked to lab notes, figures, and methodological details to enable more context-aware claim evaluation.
\citet{schlichtkrull2023intended} examine how automated fact-checking methods are framed and motivated in highly cited NLP papers, particularly in introductions, showing that claims about verification systems are often underspecified with respect to their intended use and scope. This highlights the need for clearer articulation of what different forms of verification are designed to assess. More broadly, fact checking encompasses multiple dimensions beyond binary factual correctness, including understatement, exaggeration, and contradiction \citep{kao2024we}. Our work builds on this perspective by focusing specifically on exaggeration within scientific writing, operationalising the degree to which claims are proportionally supported by evidence presented in the same paper.

In contrast to prior work on scientific claim verification, our approach targets a distinct aspect of fact-checking: assessing evidence proportionality within a single paper. Rather than determining factual correctness, we evaluate whether the strength and scope of a claim are justified by the paper’s own methods and results. We introduce a granular overstatement score to capture degrees of exaggeration, as claims are rarely contradicted by internal evidence but are often phrased more strongly than that evidence warrants. This positions our work as complementary to existing fact-checking efforts, focusing on scientific rigour and clarity rather than external verification.

\noindent\textbf{2.3~ Automatic Peer Reviewing.}~ 
Rising workloads placed on reviewers makes automating aspects of the peer-review process increasingly important \citep{staudinger2024analysis,eger2025transforming}. However, LLMs often lack the domain knowledge required to critique methodological details \citep{du2024llms}. Benchmarks such as AAAR-1.0 \citep{lou2024aaar} look into identifying paper weaknesses and reliability of reviews, and show that models can fall short in detecting subtle weaknesses.

Building on prior work showing that incorporating peer-review comments improves LLM-based evaluation accuracy \citep{zhou2024llm} and that LLMs attend selectively to different aspects of scientific feedback \citep{liang2024can}, we derive review-informed overstatement scores during annotation. This design is motivated by the observation that evaluating soundness, comparisons, and substantive claims is knowledge-intensive when relying on paper content alone. Conditioning models on reviews provides access to expert-written critiques of evidential sufficiency and overgeneralisation, yielding more consistent and calibrated overstatement judgments, consistent with prior work demonstrating the value of review text for modelling peer-review outcomes \citep{bharti2024peerrec}.

Recent ML conferences position LLMs as lightweight assistants rather than replacements in peer reviewing, supporting reproducibility checks and review quality without influencing editorial decisions \citep{neurips2025, iclr2025, AAAI2025}. In contrast to these systems, which focus on aiding the review process itself, our work leverages peer-review signals to assess whether authors’ claims are proportionally supported by the evidence presented in their papers.


\section{Data Processing Framework}

\noindent\textbf{Task Definition.}
We define two tasks for detecting overstatements in claims, focusing on claims extracted from the abstract and introduction of scientific papers: (i) \textit{Evidence Retrieval}: Given a claim, retrieve all relevant evidence that directly supports the claim. (ii) \textit{Overstatement detection}: Given a claim and its corresponding evidence, assign a continuous score indicating the degree to which the claim's wording exceeds what the evidence supports, accompanied by a brief justification.
A claim is \textit{overstated} when it makes assertions not justified by the paper’s evidence (limited experiments, lack of methodological detail, etc); \textit{partially overstated} when some components are supported but others extend beyond what the evidence warrants; and \textit{well-stated} when the claim is fully grounded in the paper’s methods, results, and reasoning without exaggeration. Examples of each claim type are provided in the case study in Table~\ref{tab:case-study-2x2}.

\subsection{Data Preparation}

We collected papers and associated reviews from OpenReview, focusing on NeurIPS and ICLR submissions. In addition to using the OpenReview API\footnote{\url{https://github.com/openreview/openreview-py}}, we incorporated previously collected ICLR datasets \citep{wang-etal-2020-reviewrobot,yuan2022can,li2023summarizing,wang2023have}.\footnote{\url{https://github.com/hughplay/ICLR2024-OpenReviewData}} 
To mitigate reviewer subjectivity and ensure consistency, we restricted our dataset to papers for which all reviewers assigned identical overall scores. Prior work has shown that reviewer disagreement is common and can substantially affect review outcomes, with repeated evaluations often leading to different accept/reject decisions \citep{beygelzimer2023has}. Review quality has also been shown to vary due to bias and miscalibration across evaluators \citep{goldberg2025peer}. Focusing on high-agreement subsets is therefore a common strategy for constructing more reliable peer-review benchmarks \citep{staudinger2024analysis,peng2025frontier}.
We processed PDFs using SciPDF\footnote{\url{https://github.com/titipata/scipdf_parser}} for text extraction and PDFFigures2 \citep{pdffigures2} for tables and figures, segmenting papers into paragraphs, figures, and tables. Due to PDF parsing failures, some papers were excluded, resulting in 659 ICLR and 213 NeurIPS papers. Reviewer scores are well distributed across the dataset (see Appendix~\ref{app:dataset_details}).

To improve evidence retrieval precision, we segment papers into smaller textual units, allowing models to operate over sentences rather than long contexts, which has been shown to improve retrieval accuracy for LLM-based reviewing tasks \citep{zhou2024llm}. Claims are extracted from the abstract and introduction, while supporting evidence is drawn from the main body, including text, tables, and figures with captions. We employ a multi-LLM annotation framework consisting of three text-only (\textsc{Seed-OSS-36B-Instruct},\textsc{gpt-oss-120b}, and \textsc{DeepSeek-R1-Distill-Qwen-32B}) and five vision-language models (VLMs) (\textsc{gemma-3-27b-it}, \textsc{Apriel-1.5-15b-Thinker}, \textsc{Kimi-VL-A3B-Instruct}, \textsc{MiniCPM-V-4\_5}, and \textsc{Qwen3-VL-30B-A3B-Instruct}), each independently annotating claims and evidence. Full model names provided in Appendix~\ref{llm_anno}. Final annotations are determined by majority vote, reducing model-specific bias and improving consistency across annotators \citep{pavlovic-poesio-2024-effectiveness,liu-etal-2024-llms-narcissistic,tseng2025evaluating,yuan2025case}.

\subsection{Author's Own Statement Extraction}

To extract claims, we define a \textit{claim} as any sentence that clearly presents an original claim, finding, or result that is central to the paper’s contributions. This definition excludes sentences that offer only background, contextual information, or references of prior work.
We first split the abstract and introduction into individual sentences using WTPsplit \citep{frohmann-etal-2024-segment},  as it consistently outperformed other sentence segmentation baselines, particularly on scientific text. For each sentence, we provided the complete abstract and introduction as context and then prompted our LLMs to classify the sentence (prompt in Table \ref{tab:prompt1} in Appendix \ref{A:prompts}). In total \textbf{10,641} claims were extracted as author's own statements. 
To assess the robustness of the annotation process, we compute Krippendorff’s $\alpha$ between the full-panel majority vote and a leave-one-out majority vote for each annotator model. Specifically, we recompute the majority label after excluding that model and measure agreement with the full-panel consensus. High agreement indicates that no single model disproportionately influences the final annotation and that the consensus labels are robust to the removal of individual annotators. Across all models, this analysis shows near-perfect agreement, suggesting stable aggregation rather than dominance by any single model (full breakdown shown in \autoref{llm_anno}).\footnote{Manual check of 200 randomly sampled sentences showed that the LLMs correctly identified over 98\% of cases.}

\subsection{Evidence Retrieval}
We segment each paper’s main body into sentences using WTPsplit and assign sentence IDs so LLMs can reference evidence by index rather than copying text, reducing hallucinations. For each claim, we prompt the LLM annotators to select supporting evidence from the paper body, prioritising results, analysis, conclusions, and other directly relevant context while excluding irrelevant or purely paraphrased content (prompt in Appendix~\ref{A:prompts}, Table~\ref{tab:prompt2}). Selected adjacent sentences are merged into coherent passages to preserve local context.
For tables and figures, LLM annotators assess whether the visual content and caption support the claim; image-based evidence is evaluated exclusively by VLMs. Following findings that retrieval quality degrades with long contexts \citep{modarressi2025nolima}, we keep annotation contexts under $\sim$1K tokens.
We further compute Krippendorff’s $\alpha$ between the full-panel majority vote and a leave-one-out majority vote for each model. Overall, agreement across models ranges from substantial to almost perfect, indicating that the aggregated annotations are stable to the removal of individual annotators (full breakdown shown in \autoref{llm_anno}).\footnote{Manual check of 100 randomly sampled pairs showed that the LLMs correctly identified over 91\% of cases.} 
Manual inspection shows that the majority of disagreements are driven by span-level variation: annotators typically agree on the core supporting evidence but differ in how much surrounding context they include. Additional variability arises from the presence of multiple valid supporting passages, with some models retrieving only a subset of the relevant evidence, while others also focus on weaker or more indirect supporting passages for the same claim.

\subsection{Overstatement Annotation}

We adopt a review-informed LLM annotation strategy for scoring claim overstatement, motivated by prior work showing that LLM-based evaluation benefits from peer-review context \citep{zhou2024llm,liang2024can}. For each claim-evidence set, the same LLM assigns a continuous overstatement score in the range $[0,1]$, where 0 denotes a well-stated claim and 1 denotes a clearly overstated claim.
We obtain multiple scores for each claim-evidence set under different annotation contexts. In a paper-only setting, the model relies solely on the paper content to assess overstatement. In review-informed settings, the same model is additionally conditioned on individual peer-review comments and produces a separate score for each review.\footnote{Peer reviews are provided as a single contextual unit, as they are written as holistic assessments whose reasoning spans multiple sentences, unlike localised evidence in the paper.} Conditioning on reviewer feedback exposes the model to expert-written critiques of evidential sufficiency and overgeneralisation, yielding judgments that more closely reflect reviewer reasoning. 
Because overstatement is inherently graded and admits legitimate disagreement, we retain all individual scores rather than aggregating them via majority voting. This produces a denser supervision signal that captures variability across models and annotation contexts. For the validation and test splits, we compute the mean score across annotations for each claim-evidence set and use this average as a soft label.

\begin{table}[!t] 
\centering
\footnotesize
\setlength{\tabcolsep}{4pt}
\renewcommand{\arraystretch}{0.8}
\begin{tabular}{lcccc}
\toprule
\textbf{Split} & \textbf{Train} & \textbf{Dev} & \textbf{Test} & \textbf{Total} \\
\midrule
Paper IDs & 536 & 259 & 77 & 872\\
Claims & 6,449 & 3,056 & 1,063 & 10,641\\
Evidence & 429,519 & 19,0414 & 62,038 & 681,971 \\
Scores & 159,930 & 3,056 & 1,063 & 164,049\\
\bottomrule
\end{tabular}
\caption{Dataset statistics. Evidence includes both supporting and not-supporting items, full breakdown shown in \autoref{params1}.}
\label{tab:dataset}
\end{table}

\noindent \textbf{Quality Control.} 
We assess the reliability of the automatically assigned overstatement scores via a human validation study with two PhD-level evaluators in Computer Science and Machine Learning. Each evaluator independently rated 30 claim-evidence sets sampled across the full score range, including both textual and visual evidence. Ratings were provided on a five-point ordinal scale (1--5), where 1 denotes a well-stated claim fully supported by evidence and 5 denotes a clearly overstated claim.
Model predictions were discretised into ordinal bins (equal-width intervals corresponding to the 1--5 human scale) and compared against human ratings using Krippendorff’s $\alpha$ (ordinal). The resulting agreement of $0.62$ indicates substantial alignment between automated and human judgments.

Incorporating peer-review context introduces a small but systematic directional shift in overstatement scores. On average, review-informed scores increase relative to paper-only scores by an average of 0.028 (median 0.005), indicating that models become modestly more critical when exposed to reviewer feedback. This shift is asymmetric: 51.0\% of scores increase after conditioning on reviews, compared to 32.6\% that decrease, while 16.4\% remain unchanged. Stratified analysis (Table~\ref{tab:review_context_bands} in Appendix) shows that reviews tend to raise scores for initially low or borderline claims, while slightly tempering scores for already high-overstatement claims, suggesting a calibration effect rather than uniform inflation. Despite these shifts, paper-only and review-informed scores remain strongly correlated (Pearson $r = 0.79$), showing that review context primarily refines existing judgments rather than overturning them, and increases the mean pairwise Pearson correlation across LLM annotators by 10.5\%, reflecting greater consistency in their relative assessments.
To assess potential annotator bias, we conducted a leave-one-model-out analysis over all claims. For each model, we recomputed the aggregated score without its annotations and compared the resulting distribution to the full baseline using Welch's $t$-test. Although several exclusions produced statistically significant differences ($p<0.01$), all absolute shifts were small (MAD $<0.03$), indicating that no single model disproportionately influences the final scores (see \autoref{llm_anno}, Table~\ref{tab:exclusion_differences}).

To prevent data leakage and assess cross-domain generalisation, we split the dataset by paper ID and exclude NeurIPS papers from the training set. Dataset statistics are summarised in Table~\ref{tab:dataset}.

\section{Experimental Setup}
\noindent \textbf{Evidence Retrieval.} 
We evaluate the learnability of intra-paper evidence retrieval using supervision derived from RIGOURATE annotations. Only text information is utilised for the task, with the captions for the tables and figures in place of the visual inputs. 
We consider a diverse set of reranker models spanning bi-encoder, cross-encoder, and generative architectures, selected based on strong performance on the MTEB benchmark.\footnote{\url{https://huggingface.co/spaces/mteb/leaderboard}} These include MiniLM \citep{reimers-2019-sentence-bert}, bge-reranker \citep{chen2024bge}, gte-reranker \citep{zhang2024mgte}, GritLM-7B \citep{muennighoff2024generative}, and Qwen3-Reranker \citep{qwen3embedding}. We additionally evaluate the E2Rank family, which combines dense retrieval with LLM-based relevance scoring \citep{liu2025e2rank}.
To assess whether the automatically constructed supervision generalises beyond the annotation process, we fine-tune the top 3 best performing model families. Full model specifications, training procedures, and prompting details are provided in \autoref{params1}.

As claims may be supported by multiple evidence items, we report MAP to assess overall ranking quality, MRR \citep{voorhees1999trec} to measure how quickly relevant evidence is retrieved, Recall@k to evaluate coverage of supporting evidence, and NDCG@k \citep{jarvelin2002cumulated} to reward placing the most informative evidence higher in the ranking.

\noindent\textbf{Overstatement Detection.} 
We selected a range of state-of-the-art text-only and VLMs spanning multiple families and sizes. Specifically, we used DeepSeek (V3.2 and R1) \cite{deepseekai2024deepseekv32}, and GLM-4.6 \cite{zeng2025glm} for text-only evaluation, and InternVL3.5 (38B and 30B-A3B) \cite{wang2025internvl3_5}, Ovis2-34B \cite{lu2024ovis}, Qwen3-VL-32B \cite{qwen3technicalreport}, GLM-4.5V \cite{vteam2025glm45vglm41vthinkingversatilemultimodal}, GPT-5-mini (low, high) \cite{openai2025gpt5} for multimodal analysis. We fine-tune several VLMs to evaluate the role of visual evidence in the task. Specifically, we selected Qwen3-VL-8B-Instruct \cite{qwen3technicalreport}, InternVL3.5-8B \cite{wang2025internvl3_5}, and LLaVA-OV-1.5-8B-Instruct \cite{an2025llavaonevision15fullyopenframework}. See \autoref{params1} for model specifications and fine-tuning details.

For evaluation, we use the concordance correlation coefficient (CCC) \citep{lawrence1989concordance}, which is well suited for overstatement detection as it measures agreement between continuous scores while penalising systematic over- or under-estimation of claim strength; values closer to 1 indicate strong agreement, while lower values reflect miscalibration or inconsistent scoring. We also report mean absolute error (MAE) to quantify the magnitude of scoring deviations and Pearson’s $\rho$ to capture relative ranking consistency independent of calibration, providing complementary views of both calibration and ordering performance.

\section{Experimental results}

\begin{table*}[t]
\centering
\footnotesize
\setlength{\tabcolsep}{5pt}
\renewcommand{\arraystretch}{0.95}

\begin{tabularx}{\textwidth}{
  >{\raggedright\arraybackslash}p{4.0cm}
  *{2}{>{\centering\arraybackslash}p{1.1cm}}
  *{3}{>{\centering\arraybackslash}p{1.1cm}}
  *{3}{>{\centering\arraybackslash}p{1.1cm}}
}
\toprule
\textbf{Model} 
& \textbf{MAP}
& \textbf{MRR}
& \textbf{R@5}
& \textbf{R@10}
& \textbf{R@20}
& \textbf{N@5} 
& \textbf{N@10}
& \textbf{N@20}
\\
\midrule

\multicolumn{9}{c}{\textit{Zero-shot models}} \\
\midrule

MiniLM-L6-v2 
& 44.08\rlap{*} & 66.12
& 10.37 & 20.46 & 39.72
& 71.48 & 71.66 & 71.82 \\

bge-reranker-v2-m3 
& 43.99 & 65.30 
& 10.35 & 20.23 & 39.73
& 70.94 & 71.25 & 71.55 \\

gte-reranker-base
& 43.99 & 65.80
& 10.47 & 20.40 & 39.43
& 71.54 & 71.62 & 71.84 \\

GritLM-7B 
& 43.15 & 66.17
& 10.16 & 19.47 & 38.22
& 70.83 & 71.39 & 71.36 \\

Qwen3-Reranker-0.6B 
& 44.19 & 66.73
& 10.54 & 20.71 & 39.77
& 72.07 & 72.05 & 72.14 \\

Qwen3-Reranker-4B 
& 44.92 & 68.74
& 10.93 & 21.02 & 40.71
& 73.37 & 73.13 & 73.12 \\

Qwen3-Reranker-8B 
& 45.88\rlap{*} & 71.72
& 11.09 & 21.52 & 41.49
& 75.34 & 74.65 & 74.28 \\

E2Rank-0.6B
& 47.57\rlap{*} & 85.54
& 12.86 & 22.80 & 41.18
& 85.19 & 81.75 & 79.20 \\

E2Rank-4B
& 47.41 & 86.06
& 12.67 & 22.68 & 41.07
& 85.16 & 81.63 & 79.13 \\

E2Rank-8B
& 47.54 & 85.77
& 12.57 & 22.78 & 41.36
& 85.36 & 81.80 & 79.25 \\
\midrule
\multicolumn{9}{c}{\textit{Fine-tuned (Top-3 models only)}} \\
\midrule

MiniLM-L6-v2
& 45.51 {\color{mygreen}\scriptsize(+1.4)}
& 74.62 {\color{mygreen}\scriptsize(+8.5)}
& 11.33 {\color{mygreen}\scriptsize(+1.0)}
& 21.37 {\color{mygreen}\scriptsize(+0.9)}
& 40.00 {\color{mygreen}\scriptsize(+0.3)}
& 77.41 {\color{mygreen}\scriptsize(+5.9)}
& 76.14 {\color{mygreen}\scriptsize(+4.5)}
& 75.02 {\color{mygreen}\scriptsize(+3.2)} \\

Qwen3-Reranker-8B 
& \textbf{54.19} {\color{mygreen}\scriptsize(+8.3)}
& \underline{83.44} {\color{mygreen}\scriptsize(+11.7)}
& \textbf{14.84} {\color{mygreen}\scriptsize(+3.8)}
& \textbf{27.32} {\color{mygreen}\scriptsize(+5.8)}
& \textbf{48.26} {\color{mygreen}\scriptsize(+6.8)}
& \underline{85.19} {\color{mygreen}\scriptsize(+9.9)}
& \underline{83.33} {\color{mygreen}\scriptsize(+8.7)}
& \textbf{82.11} {\color{mygreen}\scriptsize(+7.8)} \\

E2Rank-0.6B
& \underline{52.37} {\color{mygreen}\scriptsize(+4.8)}
& \textbf{86.57} {\color{mygreen}\scriptsize(+0.4)}
& \underline{14.69} {\color{mygreen}\scriptsize(+1.2)}
& \underline{25.96} {\color{mygreen}\scriptsize(+3.2)}
& \underline{46.34} {\color{mygreen}\scriptsize(+5.1)}
& \textbf{86.75} {\color{mygreen}\scriptsize(+1.6)}
& \textbf{83.95} {\color{mygreen}\scriptsize(+2.2)}
& \underline{82.06} {\color{mygreen}\scriptsize(+2.1)} \\
\bottomrule

\end{tabularx}

\caption{
Task 1: Retrieval performance using MAP, MRR, Precision@5/10/20, and NDCG@5/10/20 (N).  
Fine-tuning for the top-3 rerankers (MiniLM-L6-v2, Qwen3-Reranker-8B, and E2Rank-0.6B), indicated by the * next to MAP scores. Green values in brackets indicate relative gains against the Base setting.
\textbf{Bold} = best; \underline{underline} = 2nd best.
}
\label{tab:evidence_retrieval}
\end{table*}

\begin{table}[!tb]
\centering
\footnotesize
\setlength{\tabcolsep}{6pt}
\renewcommand{\arraystretch}{0.91}
\begin{tabular}{lccc}
\toprule
\textbf{Model} & \textbf{CCC $\uparrow$} & \textbf{MAE $\downarrow$} & \textbf{$\rho\uparrow$} \\
\midrule
\multicolumn{4}{c}{\textit{Text-only models}} \\
\midrule
Deepseek-V3.2 & 0.356 & 0.195 & 0.392\\
Deepseek-R1 & 0.463 & 0.201 & 0.544\\
GLM-4.6 & 0.385 & 0.240 & 0.490\\
\midrule
\multicolumn{4}{c}{\textit{Vision–language models (VLMs)}} \\
\midrule
InternVL3.5-8B & 0.106 & 0.326 & 0.158 \\
InternVL3.5-38B & 0.347 & \underline{0.161} & 0.360 \\
InternVL3.5-30B-A3B & 0.133 & 0.295 & 0.257 \\
\noalign{\vskip 1mm}
\hdashline
\noalign{\vskip 1mm}
Qwen3-VL-8B & 0.323 & 0.237 & 0.428 \\ 
Qwen3-VL-32B & 0.456 & 0.187 & 0.532\\
\noalign{\vskip 1mm}
\hdashline
\noalign{\vskip 1mm}
GPT-5-mini (low) & \underline{0.478} & 0.209 & \underline{0.571}\\
GPT-5-mini (high) & \textbf{0.493} & 0.204 & \textbf{0.587}\\
\noalign{\vskip 1mm}
\hdashline
\noalign{\vskip 1mm}
LLaVA-OV-1.5-8B  & 0.088 & 0.241 & 0.116 \\
Ovis2-34B  & \textbf{0.493} & \textbf{0.154} & 0.509 \\ 
GLM-4.5V & 0.358 & 0.169 & 0.447\\
\bottomrule
\end{tabular}
\caption{Base model performance on overstatement detection utilising the claim-evidence sets.Higher is better for CCC and $\rho$, lower is better for MAE; \textbf{Bold} = best; \underline{underline} = 2nd best.}
\label{tab:base_model_metrics}
\end{table}

\begin{table}[!tb]
\centering
\footnotesize
\setlength{\tabcolsep}{3.9pt}
\renewcommand{\arraystretch}{0.92}
\begin{tabularx}{\linewidth}{l l c c c}
\toprule
\textbf{Model} & \textbf{Setting} & \textbf{CCC $\uparrow$} & \textbf{MAE$\downarrow$} & $\boldsymbol{\rho\uparrow}$ \\
\midrule
Qwen3-VL-8B      & Base       & 0.323 & 0.237 & 0.428 \\
                 & Text-only  & \underline{0.529} & \underline{0.156} & \underline{0.593} \\
                 & +Image     & \textbf{0.578} & \textbf{0.153} & \textbf{0.649} \\
\midrule
\addlinespace[2pt]
InternVL3.5-8B   & Base       & 0.106 & 0.326 & 0.158 \\
                 & Text-only  & 0.418 & 0.208 & 0.528 \\
                 & +Image     & 0.479 & 0.191 & 0.557 \\
\midrule
\addlinespace[2pt]
LLaVA-OV-1.5-8B  & Base       & 0.088 & 0.241 & 0.116 \\
                 & Text-only  & 0.230 & 0.229 & 0.326 \\
                 & +Image     & 0.317 & 0.205 & 0.392 \\
\bottomrule
\end{tabularx}
\caption{
Performance before and after fine-tuning under three settings: Base (zero-shot), Text-only (fine-tuned using text inputs only), and +Image (fine-tuned with visual and text inputs).
Higher is better for CCC and $\rho$, lower is better for MAE; \textbf{Bold} = best; \underline{underline} = 2nd best.
}
\label{tab:finetuned_vs_base}
\end{table}

\noindent\textbf{Evidence Retrieval.}~In the zero-shot setting, rerankers with generative or hybrid relevance modelling consistently outperform encoder-only models, suggesting that matching scientific claims to internal evidence benefits from richer semantic reasoning beyond embedding similarity. Among zero-shot methods, the E2Rank family performs strongly across metrics, with E2Rank-0.6B achieving competitive MAP and recall despite its smaller size, indicating that hybrid embedding–reranking approaches are effective for intra-paper evidence selection (as shown in Table \ref{tab:evidence_retrieval}).

Fine-tuning leads to substantial and consistent improvements across all evaluated models, confirming that evidence retrieval within scientific papers is a learnable task under the proposed automatic supervision. In particular, Qwen3-Reranker-8B exhibits the largest gains across MAP, Recall@K, and NDCG@K, suggesting that conventional reranking architectures are able to exploit the task-specific supervision.

\noindent\textbf{Overstatement Detection.}~
Table~\ref{tab:base_model_metrics} shows that strong text-only models can achieve competitive performance on overstatement detection, with DeepSeek-R1 substantially outperforming other text-only baselines and performing on par with several VLMs. This indicates that linguistic cues alone capture part of the signal, particularly for claims whose evidential support is primarily textual. VLMs such as GPT-5-mini (high) and Ovis2-34B achieve the strongest overall performance, exhibiting both higher agreement (CCC) and more reliable ranking (Pearson’s $\rho$), while also maintaining lower MAE. Although absolute CCC values are modest, scores around 0.5 are expected for graded, subjective judgments and correspond to strong agreement given the metric’s sensitivity to calibration. The variance among VLMs further suggests that current gains from visual inputs may be constrained by limitations in multimodal reasoning.

Table~\ref{tab:finetuned_vs_base} analyses the effect of fine-tuning and visual grounding within the same model families. Fine-tuning substantially improves agreement and ranking consistency (CCC and Pearson’s $\rho$), confirming that overstatement detection is learnable under the proposed supervision. Incorporating visual inputs further increases CCC and $\rho$ across models, indicating that figures and tables provide complementary signal for assessing relative overstatement severity. Improvements in absolute error (MAE) are more model-dependent, particularly for InternVL and LLaVA; this is expected since reference scores are soft labels obtained by averaging multiple annotations, where gains are more reliably reflected in agreement and ranking than in exact score matching. Overall, multimodal fine-tuning enables smaller models to match or exceed the performance of much larger zero-shot counterparts.

\begin{table*}[t]
\centering
\begin{tabular}{@{}p{0.47\textwidth}@{\hspace{0.04\textwidth}}p{0.47\textwidth}@{}}

\fbox{\parbox[t]{\linewidth}{%
\vspace{1mm}
\footnotesize\textbf{Claim:} 
Our analyses indicate a 
\textcolor{orange}{\textbf{consistent inward movement}} 
of embeddings accompanied by 
\textcolor{blue}{\textbf{reduced anisotropy}}, 
which \textcolor{teal}{\textbf{clarifies under-represented regions}}.

\vspace{2mm}
\hrule
\vspace{1mm}

\textbf{Score:} \textbf{0.6}

\vspace{1mm}

\textbf{Reasoning:} 
The claim uses strong language ("consistent," "clarifies") that isn't fully supported by the evidence. The evidence shows a trend of inward movement and reduced anisotropy, but doesn't demonstrate a complete or consistent effect. The claim about clarifying under-represented regions is not directly supported by the provided data, which focuses on geometric changes and isotropy rather than semantic coverage. The evidence also lacks details on the specific regions being clarified, making the claim an overstatement of the findings.

\vspace{1mm}
}}
&
\fbox{\parbox[t]{\linewidth}{%
\vspace{1mm}
\footnotesize\textbf{Claim:}
We \textcolor{orange}{\textbf{formally prove}} 
that contrastive learning induces embeddings to 
\textcolor{blue}{\emph{\textbf{converge at the origin}}}.

\vspace{5mm}
\hrule
\vspace{1mm}

\textbf{Score:} \textbf{0.85}

\vspace{1mm}

\textbf{Reasoning:}
The claim uses the strong term "formally prove," which implies a rigorous mathematical proof, but the evidence consists of empirical observations, visualizations, and correlations (e.g., L2-norm trends, UMAP plots) without any formal proof. The evidence shows that embeddings tend to move toward the origin and become more isotropic, but this does not constitute a formal proof of convergence. The claim overstates the strength of the evidence by implying a level of theoretical certainty that is not supported by the data.

\vspace{5mm}
}}
\\[0mm]


\fbox{\parbox[t]{\linewidth}{%
\vspace{1mm}
\footnotesize\textbf{Claim:} 
We find that the embedding space is 
\textcolor{blue}{\textbf{directed towards the origin}} during training,
with 
\textcolor{teal}{\textbf{more areas now better defined}}.

\vspace{2mm}
\hrule
\vspace{1mm}

\textbf{Score:} \textbf{0.2}

\vspace{1mm}

\textbf{Reasoning:} 
The claim accurately states that embeddings move toward the origin and become more defined, supported by evidence showing reduced anisotropy and better-defined areas.However, it slightly overstates by not mentioning the role of temperature and the specific conditions under which these changes occur, leading to minor exaggeration.

\vspace{8mm}
\vspace{1mm}
}}
&

\fbox{\parbox[t]{\linewidth}{%
\vspace{1mm}
\footnotesize\textbf{Claim:} 
Our \textcolor{orange}{\textbf{observations suggest}} that the movement of embeddings toward the origin is 
\textcolor{blue}{\textbf{one of several dynamics}} 
contributing to the organization of the semantic space.

\vspace{2mm}
\hrule
\vspace{1mm}

\textbf{Score:} \textbf{0.1}

\vspace{1mm}

\textbf{Reasoning:} 
The claim is well-supported by the evidence, which demonstrates that embedding movement toward the origin is one of several dynamics contributing to semantic space organization. The evidence includes experiments showing the effects of temperature and pooling methods on isotropy, as well as visualizations of embedding changes. The claim does not overstate the evidence, as it accurately reflects the findings without exaggeration.

\vspace{1mm}
}}
\\

\end{tabular}

\caption{A case study demonstrating different degrees of overstatement. Colour-coding highlights the key phrases in each claim that influence the model’s judgement, and the full supporting evidence is provided in \autoref{app:case_study}.}
\label{tab:case-study-2x2}
\end{table*}

\section{Case Study}\label{sec:case_study}

To illustrate how our framework behaves in practice, we conduct a case study on ``\textit{On Isotropy, Contextualization and Learning Dynamics of Contrastive-based Sentence Representation Learning}'' \cite{xiao-etal-2023-isotropy}, with permission from the authors. Starting from a claim in the paper, we manually construct additional variants, with varying degrees of overstatement, by modifying only the claim’s wording while keeping the supporting evidence fixed. This controlled manipulation allows us to directly probe whether the framework is sensitive to rhetorical amplification rather than differences in evidence. We utilise our best fine-tuned Qwen3-8B model, whose performance is reported in Table \ref{tab:finetuned_vs_base}, for our case study.

Table~\ref{tab:case-study-2x2} shows a clear progression in overstatement scores. The overstated variant employs absolute and theoretical language (e.g., ``consistent'', ``formally prove'') and introduces effects not directly measured in the experiments, resulting in the highest score. The partially overstated claim blends valid observations with unsupported generalisations, while the well-stated variant adheres closely to the reported results, using cautious language (e.g., ``observations suggest'') that matches the strength of the evidence. Overall, this case study highlights the core distinction targeted by our task: overstatement is not about factual incorrectness, but about how far a claim’s wording stretches beyond its evidential footing. This observation is further supported by an analysis of linguistic certainty across different degrees of overstatement, as shown in \autoref{app:certainty}. Although illustrative, this case study provides a concrete example of the framework’s behaviour in practice and complements the broader quantitative results reported in the previous sections.

\section{Conclusion}
We present a framework for detecting scientific overstatement by assessing whether the strength of claims in abstracts and introductions is proportionate to the evidence provided in the paper. By aligning claims with multimodal evidence and incorporating peer-review context during annotation, we capture graded differences in evidential support rather than binary factual correctness. Our experiments show that overstatement detection is learnable from these annotations and benefits from multimodal grounding, while also highlighting current limitations in model reasoning over visual evidence. Overall, this work reframes scientific rigour as a question of evidential proportionality and provides a foundation for tools that support clearer and more faithful scientific communication.

\section*{Limitations}
Overall, the model produces explanations that align with the authors’ own assessment. While our work is intended to encourage clearer, well grounded scientific communication, it could be misused. It is not intended to replace peer reviewing but to assist in the process. A high overstatement score should therefore be treated as a prompt for closer human review, not a basis for rejection.

Our approach is designed around the structure of scientific papers available in OpenReview, and naturally reflects the conventions and formatting typically found in this setting. As a result, the system may require adaptation when applied to other venues or scientific fields with different writing styles or evidence formats. In addition, our evaluations focus on alignment between claims and the evidence presented within a paper, rather than on broader scientific correctness, so the tool should be viewed as supporting clarity and grounding rather than making judgments about the overall quality of the work.

\section*{Acknowledgements}
Joseph James is supported by the Centre for Doctoral Training in Speech and Language Technologies (SLT) and their Applications funded by UK Research and Innovation [grant number EP/S023062/1]. We acknowledge IT Services at The University of Sheffield for the provision of services for High Performance Computing.

\bibliography{custom, anthology}
\appendix

\section{LLM annotation}\label{llm_anno}
\paragraph{Model selection}
All models were obtained from Hugging Face. Specifically, we used \textsc{ByteDance-Seed/Seed-OSS-36B-Instruct}, \textsc{openai/gpt-oss-120b}, \textsc{google/gemma-3-27b-it}, \textsc{ServiceNow-AI/Apriel-1.5-15b-Thinker}, \textsc{deepseek-ai/DeepSeek-R1-Distill-Qwen-32B}, \textsc{moonshotai/Kimi-VL-A3B-Instruct}, \textsc{openbmb/MiniCPM-V-4\_5}, and \textsc{Qwen/Qwen3-VL-30B-A3B-Instruct}. 
These models span diverse architectures and training paradigms, covering both text-only and vision–language reasoning models.

Tables~\ref{tab:annotator_alpha_modes} and~\ref{tab:exclusion_differences} report analyses assessing the robustness of the annotation and aggregation procedure under leave-one-model-out settings, while Table~\ref{tab:review_context_bands} summarises the effect of incorporating peer-review context across different initial overstatement levels.

\begin{table}[H]
\centering
\setlength{\tabcolsep}{3pt}
\renewcommand{\arraystretch}{0.88}
\begin{tabular}{@{}lccc@{}}
\toprule
\textbf{Excluded Model} & \textbf{Own} & \textbf{Text} & \textbf{Image} \\
\midrule
Seed-OSS-36B         & 0.9837 & 0.8421 & — \\
GPT-OSS-120B         & 0.9801 & 0.7746 & — \\
DeepSeek-R1-32B      & 0.9801 & 0.8368 & — \\
Qwen3-VL-30B-A3B     & 0.9806 & 0.7536 & 0.7531 \\
Gemma-3-27B-it       & 0.9822 & 0.8968 & 0.7553 \\
Kimi-VL-A3B          & 0.9815 & 0.8597 & 0.9268 \\
Apriel-1.5-15B       & 0.9892 & 0.9950 & 0.8497 \\
MiniCPM-V-4.5       & 0.9823 & 0.9945 & 0.7526 \\
\bottomrule
\end{tabular}
\caption{Krippendorff’s $\alpha$ agreement between full and excluded-model consensus across annotation settings (all $p<0.01$). 
\textbf{Own:} Authors original statement annotation. 
\textbf{Text:} Relevant text annotation. 
\textbf{Image:} Relevant visual annotation.}
\label{tab:annotator_alpha_modes}
\end{table}

\begin{table}[H]
\centering
\small
\begin{tabular}{lrrc}
\toprule
\textbf{Model} &
\textbf{$\Delta$ Mean} &
\textbf{MAD} &
\textbf{Welch $p$} \\
\midrule
Seed-OSS-36B         & +0.0160 & 0.0305 & $<0.01$ \\
GPT-OSS-120B         & +0.0085 & 0.0239 & $<0.01$ \\
Gemma-3-27B-it       & $-$0.0191 & 0.0203 & $<0.01$ \\
Apriel-1.5-15B       & +0.0024 & 0.0182 & 0.3167 \\
DeepSeek-R1-32B      & +0.0105 & 0.0244 & $<0.01$ \\
MiniCPM-V-4.5        & $-$0.0110 & 0.0306 & $<0.01$ \\
Qwen3-VL-30B-A3B     & $-$0.0074 & 0.0280 & $<0.01$ \\
Kimi-VL-A3B          & $-$0.0035 & 0.0240 & 0.084 \\
\bottomrule
\end{tabular}
\vspace{2mm}
\caption{
Change in mean overstatement scores when excluding each annotator model. $\Delta$~Mean measures how much the overall average score shifts relative to the full-model baseline, while MAD (Mean Absolute Deviation) reflects the average absolute per-claim change in the aggregated scores.
}
\label{tab:exclusion_differences}
\end{table}

\begin{table}[H]
\centering
\scriptsize
\setlength{\tabcolsep}{3pt}
\renewcommand{\arraystretch}{1.05}
\begin{tabular}{lcccc}
\toprule
\textbf{Initial score band} &
$\Delta\mu$ &
$\tilde{\Delta}$ &
$|\Delta|$ &
$\uparrow/\downarrow/=$ (\%) \\
\midrule
Low (0.0--0.3)      & +0.0978 & +0.0625 & 0.1157 & 73.0 / 16.2 / 10.8 \\
Low--Mid (0.3--0.5) & +0.0471 & +0.0333 & 0.1107 & 58.1 / 34.9 / 7.0 \\
Mid (0.5--0.7)      & -0.0457 & +0.0000 & 0.0696 & 22.3 / 41.1 / 36.7 \\
High (0.7--1.0)     & -0.1362 & -0.0889 & 0.1475 & 11.7 / 79.3 / 9.0 \\
\bottomrule
\end{tabular}
\vspace{1mm}
\caption{
Impact of peer-review context stratified by initial (paper-only) overstatement score.
Positive values indicate increased criticality after incorporating reviews, while negative values indicate reduced scores.
}
\label{tab:review_context_bands}
\end{table}

\section{Evidence retrieval: Training details}\label{params1}

Full evidence retrieval dataset breakdown shown in Table \ref{tab:evidence_breakdown}. All models were tested and trained on a single A100 GPU, with hyperparameters provided in Table~\ref{tab:training-details_evidence} Prompt used for LLM-based models is show in Table \ref{tab:finetune_prompt_task1}. Full training details provided in Table \ref{tab:training-details_evidence}.

\begin{table}[!htb]
\centering
\footnotesize
\setlength{\tabcolsep}{4pt}
\renewcommand{\arraystretch}{0.8}
\begin{tabular}{lcccc}
\toprule
\textbf{Evidence Type} & \textbf{Train} & \textbf{Dev} & \textbf{Test} & \textbf{Total} \\
\midrule
Supporting & 183,518 & 77,333 & 24,548 & 285,399 \\
TEXT & 146,243 & 62,724 & 20,693 & 229,660 \\
IMAGE & 37,275 & 14,609 & 3,855 & 55,739 \\
\midrule
Not-supporting & 246,001 & 113,081 & 37,490 & 396,572 \\
TEXT & 195,942 & 91,955 & 31,815 & 319,712 \\
IMAGE & 50,059 & 21,126 & 5,675 & 76,860 \\
\bottomrule
\end{tabular}
\caption{Detailed breakdown of evidence.}
\label{tab:evidence_breakdown}
\end{table}

\begin{table}[H]
\centering
\footnotesize
\begin{tabular}{p{0.95\linewidth}} 
\toprule
\texttt{You are an evidence verification assistant. Given a claim and a document, determine if the document provides supporting evidence for the claim.} \\
\texttt{\textbf{INSTRUCTION: }}\texttt{Does the following document provide supporting evidence for the claim?} \\
\bottomrule
\end{tabular}
\caption{Prompt for finetuning LLM based reranker models.}
\label{tab:finetune_prompt_task1}
\end{table}

\begin{table}[ht]
\centering
\begin{tabularx}{0.85\linewidth}{l l}
\toprule
\textbf{Hyperparameter}           &  \\
\midrule
epochs                            & 3 \\
batch\_size                       & 16 \\
gradient\_accumulation\_steps     & 4 \\
gradient\_checkpointing           & True \\
optim                             & adamw \\
learning\_rate                    & 3e-5 \\
weight\_decay                     & 0.01 \\
max\_grad\_norm                   & 1.0 \\
warmup\_ratio                     & 0.1 \\
logging\_steps                    & 500 \\
eval\_steps                       & 1000 \\
eval\_strategy                    & steps \\
save\_steps                       & 1000 \\
EarlyStoppingCallback             & 5 \\
\bottomrule
\end{tabularx}
\caption{Training hyperparameters for evidence retrieval.}
\label{tab:training-details_evidence}
\end{table}

\paragraph{Model selection}
All models were obtained from Hugging Face. Specifically, we used \textsc{Alibaba-NLP/E2Rank-(0.6B,4B,8B)}, \textsc{Qwen/Qwen3-Reranker-(0.6B,4B,8B)}, \textsc{sentence-transformers/all-MiniLM-L6-v2}, \textsc{BAAI/bge-reranker-v2-m3}, \textsc{Alibaba-NLP/gte-multilingual-reranker-base}, and \textsc{GritLM/GritLM-7B}.

\section{Overstatement detection: Training details}\label{params}
All models were tested and trained on a single A100 GPU, with hyperparameters provided in Table~\ref{tab:training-details_overstatement}. Prompt used for fine-tuning Task 2 is shown in Table \ref{tab:finetune_prompt}. Full training details provided in Table \ref{tab:training-details_overstatement}.

\begin{table}[H]
\centering
\footnotesize
\begin{tabular}{p{0.95\linewidth}} 
\toprule
\texttt{You are a model specialized in assessing overstated claims using text and image evidence.} \\
\texttt{You must score each claim based on how overstated or exaggerated it is with respect to the evidence,on a continuous scale from 0 to 1 where 0 means well-stated and 1 means overstated.} \\
\texttt{Provide the final score as <score>value</score> followed by a brief reasoning.} \\
\bottomrule
\end{tabular}
\caption{Prompt for finetuning for overstatement detection.}
\label{tab:finetune_prompt}
\end{table}

\begin{table}[ht]
\centering
\begin{tabularx}{0.85\linewidth}{l l}
\toprule
\textbf{Hyperparameter}           &  \\
\midrule
epochs                            & 3 \\
batch\_size                       & 1 \\
gradient\_accumulation\_steps     & 16 \\
gradient\_checkpointing           & True \\
optim                             & adamw \\
learning\_rate                    & 3e-5 \\
weight\_decay                     & 0.01 \\
max\_grad\_norm                   & 1.0 \\
warmup\_ratio                     & 0.1 \\
logging\_steps                    & 500 \\
eval\_steps                       & 1000 \\
eval\_strategy                    & steps \\
save\_steps                       & 1000 \\
EarlyStoppingCallback             & 5 \\
\bottomrule
\end{tabularx}
\caption{Training hyperparameters for overstatement detection.}
\label{tab:training-details_overstatement}
\end{table}

\paragraph{Model selection}

For larger open-source and closed-source models (\textsc{zai-org/GLM-4.5V}, \textsc{zai-org/GLM-4.6}, \textsc{DeepSeek-V3.2}. and \textsc{gpt-5-mini-2025-08-07}), we utilised APIs to obtain model outputs through their hosted inference endpoints, as local deployment was not feasible due to GPU memory limitations. For all other models we utilised a single A100 for the following models; \textsc{Qwen/Qwen3-VL-32B-Instruct}, \textsc{OpenGVLab/InternVL3\_5-38B-HF}, \textsc{OpenGVLab/InternVL3\_5-30B-A3B-HF}, and \textsc{AIDC-AI/Ovis2-34B}.
For finetuning we utilised \textsc{lmms-lab/LLaVA-OneVision-1.5-8B-Instruct}, \textsc{OpenGVLab/InternVL3\_5-4B-HF}, and \textsc{Qwen/Qwen3-VL-8B-Instruct}.

\section{Human evaluation}\label{A:human_eval}
Human evaluators were provided 20GBP in Amazon gift cards per hour to complete the evaluation, which is above the minimum wage in the UK. Further allowing for up to 3 hours of work to allow time for a thorough analysis of the claim and evidence.

\section{Dataset details}\label{app:dataset_details}
\begin{figure}[t]
\centering
\includegraphics[width=0.9\linewidth]{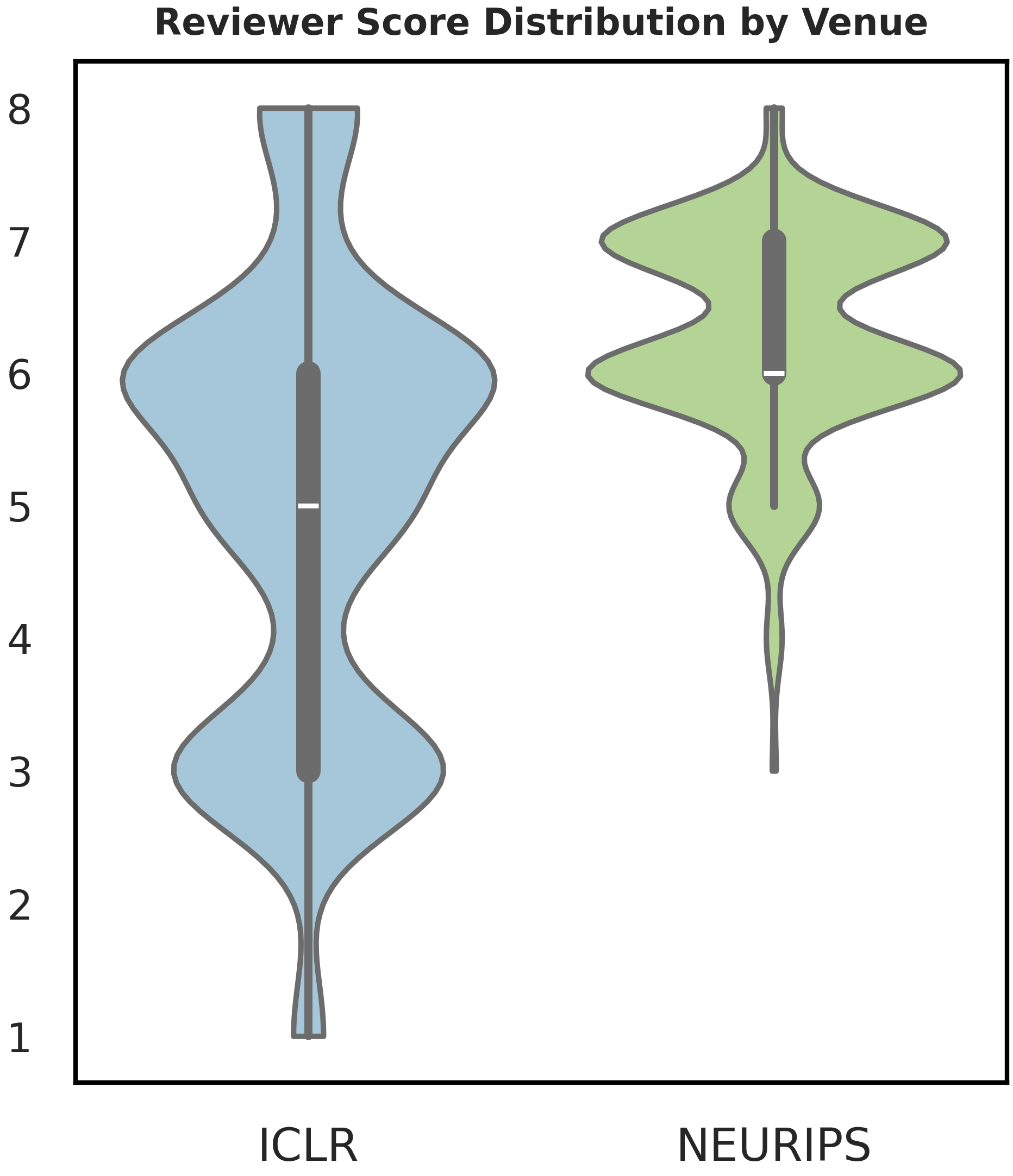}
\caption{Reviewer rating distribution for ICLR and NeurIPS.}
\label{fig:data_dist}
\end{figure}

\begin{figure}[t]
\centering
\includegraphics[width=0.9\linewidth]{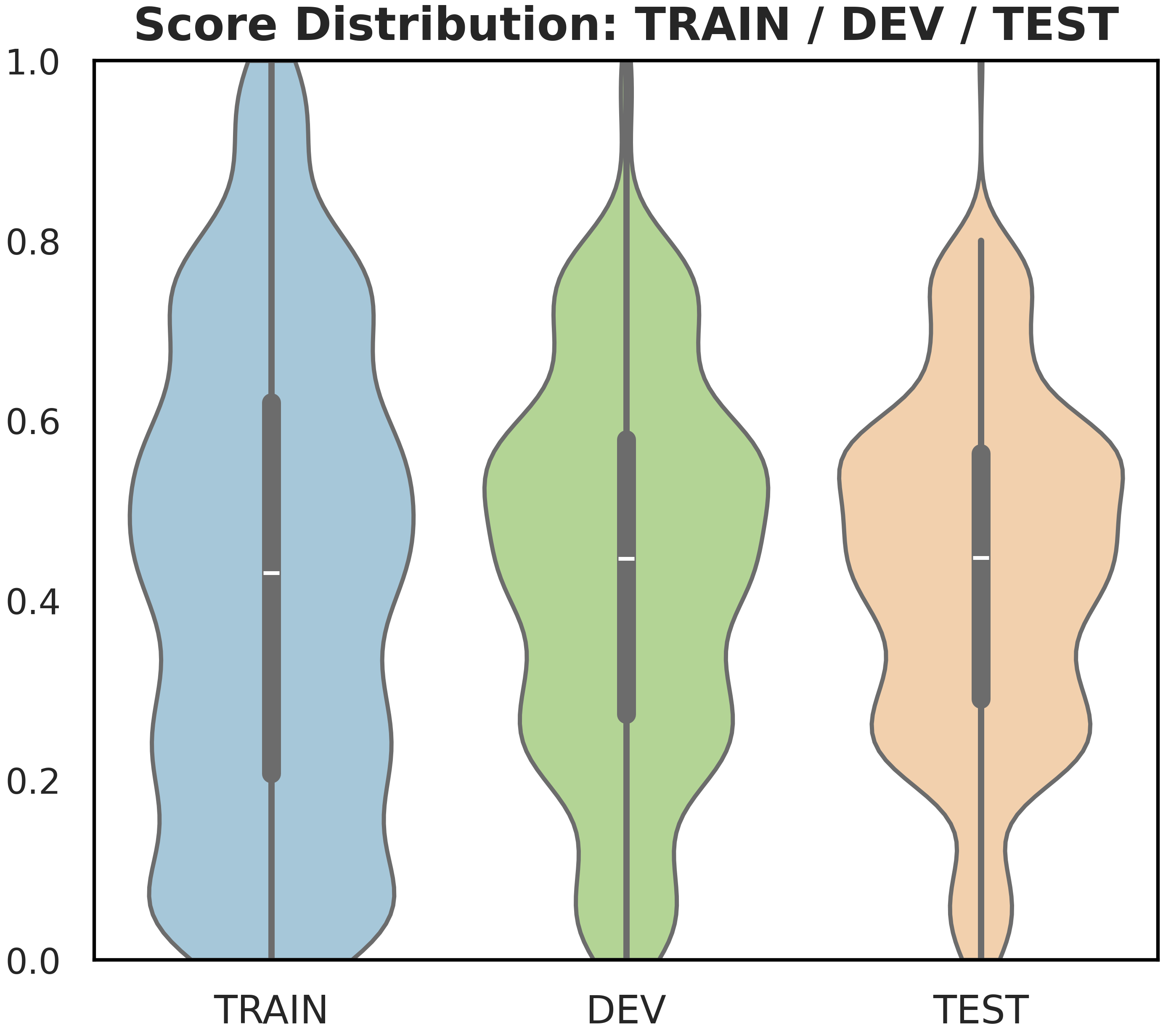}
\caption{Score distributions for each split}
\label{fig:data_overstatement_dist}
\end{figure}


\section{Certainty Measure}\label{app:certainty}

\begin{figure*}[t]
\centering

\begin{subfigure}{0.45\textwidth}
    \centering
    \includegraphics[width=\linewidth]{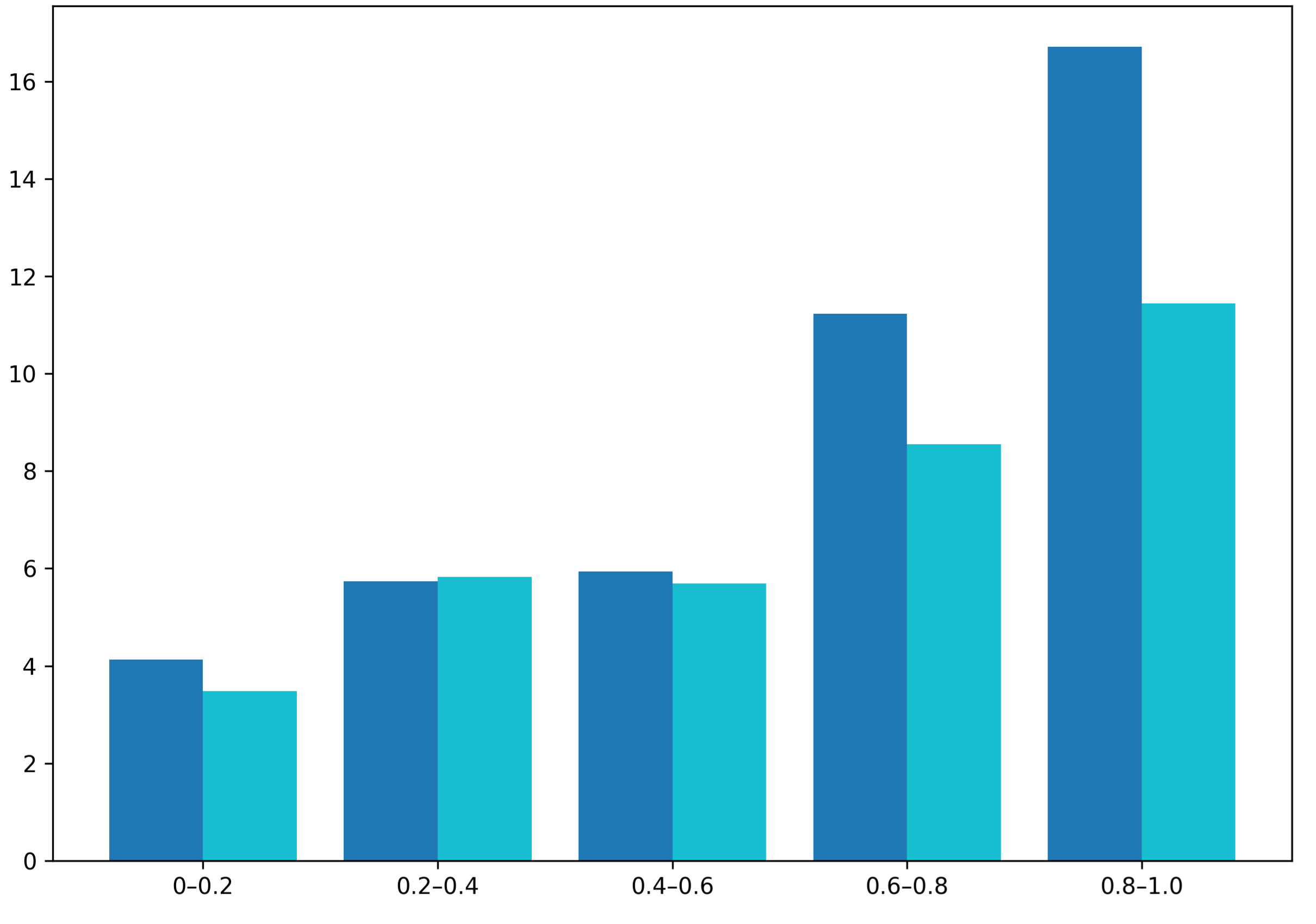}
    \caption{Extent}
\end{subfigure}
\hfill
\begin{subfigure}{0.45\textwidth}
    \centering
    \includegraphics[width=\linewidth]{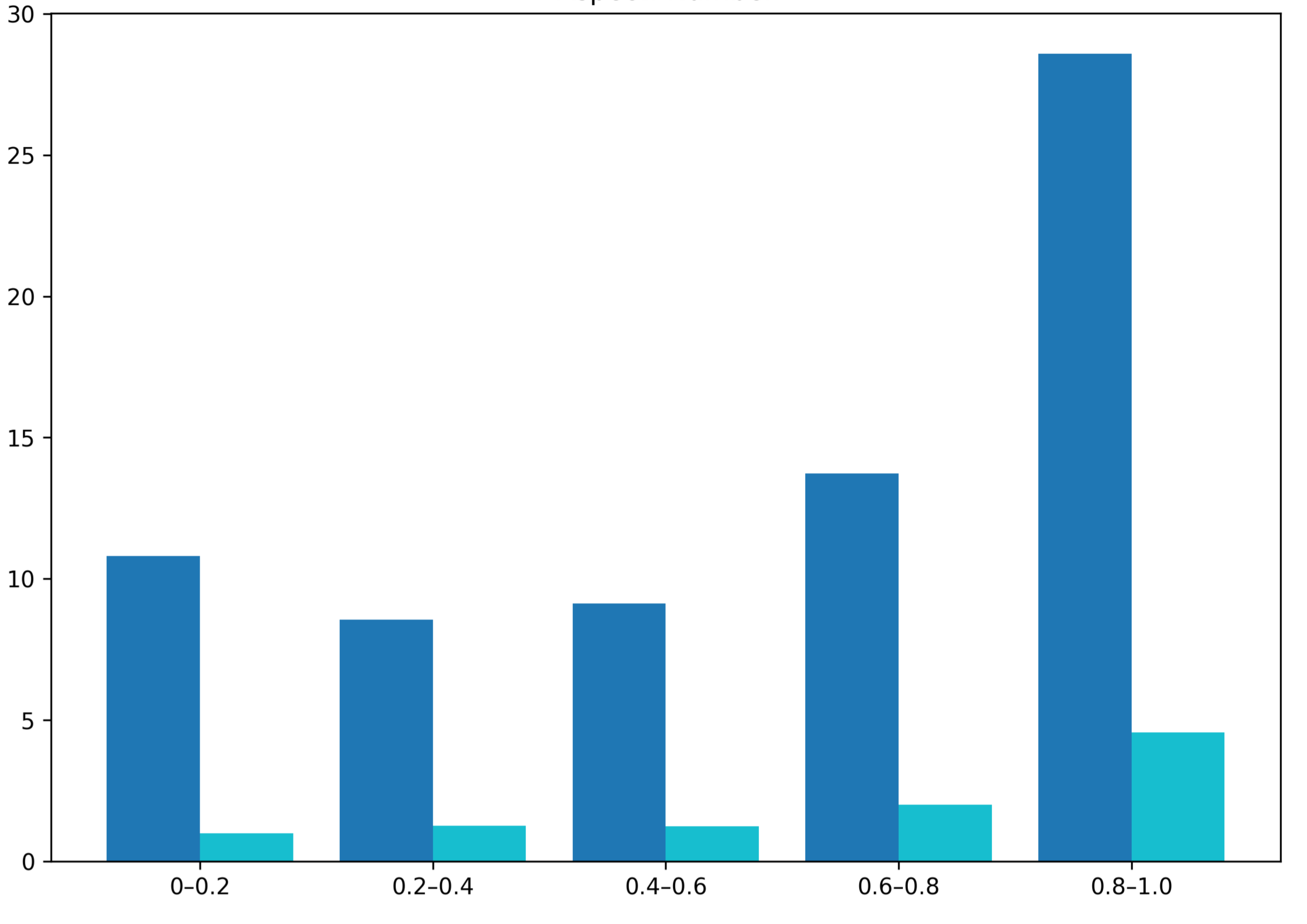}
    \caption{Number}
\end{subfigure}

\vspace{0.8em}

\begin{subfigure}{0.45\textwidth}
    \centering
    \includegraphics[width=\linewidth]{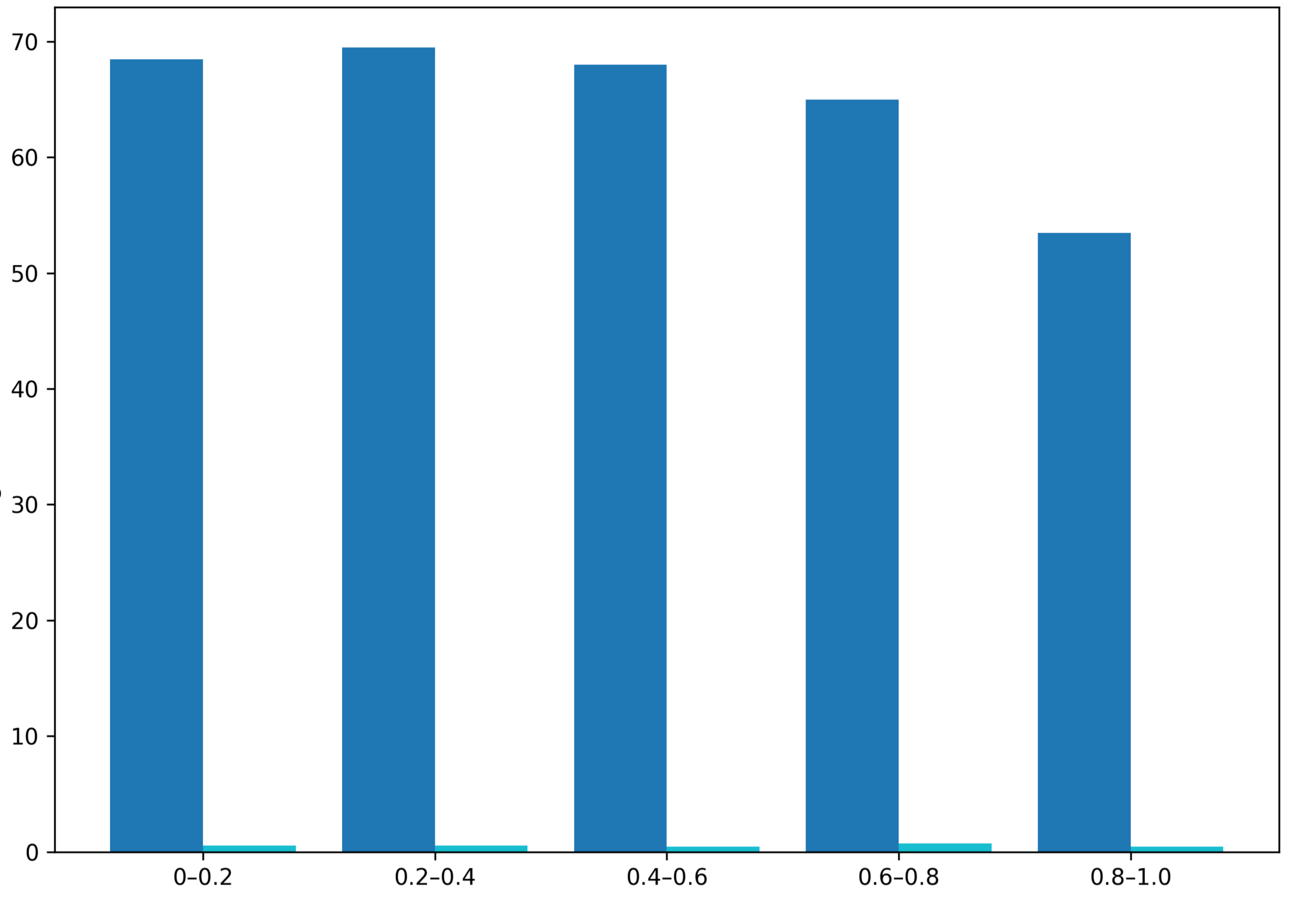}
    \caption{Framing}
\end{subfigure}
\hfill
\begin{subfigure}{0.45\textwidth}
    \centering
    \includegraphics[width=\linewidth]{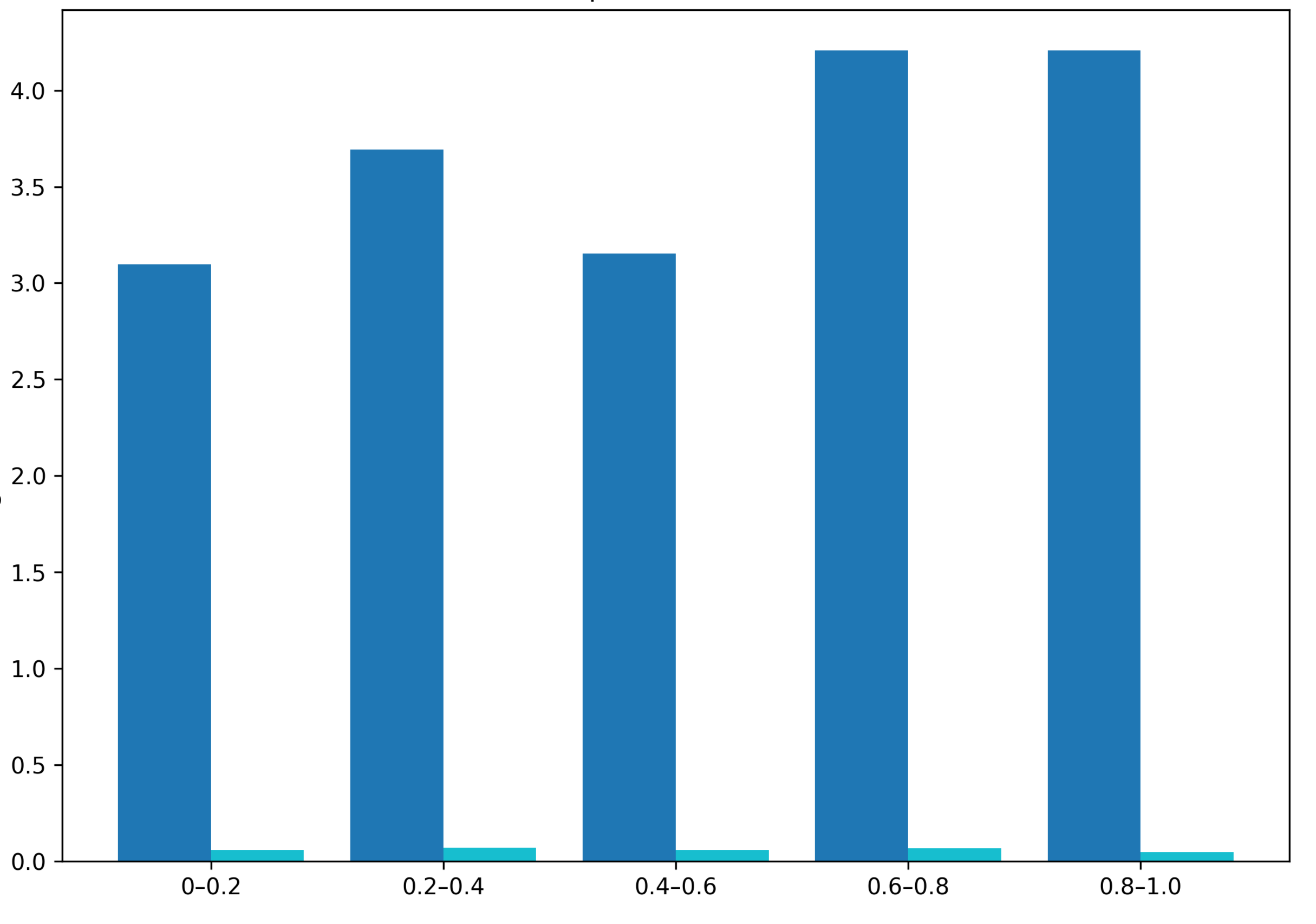}
    \caption{Condition}
\end{subfigure}

\vspace{0.8em}

\begin{subfigure}{0.45\textwidth}
    \centering
    \includegraphics[width=\linewidth]{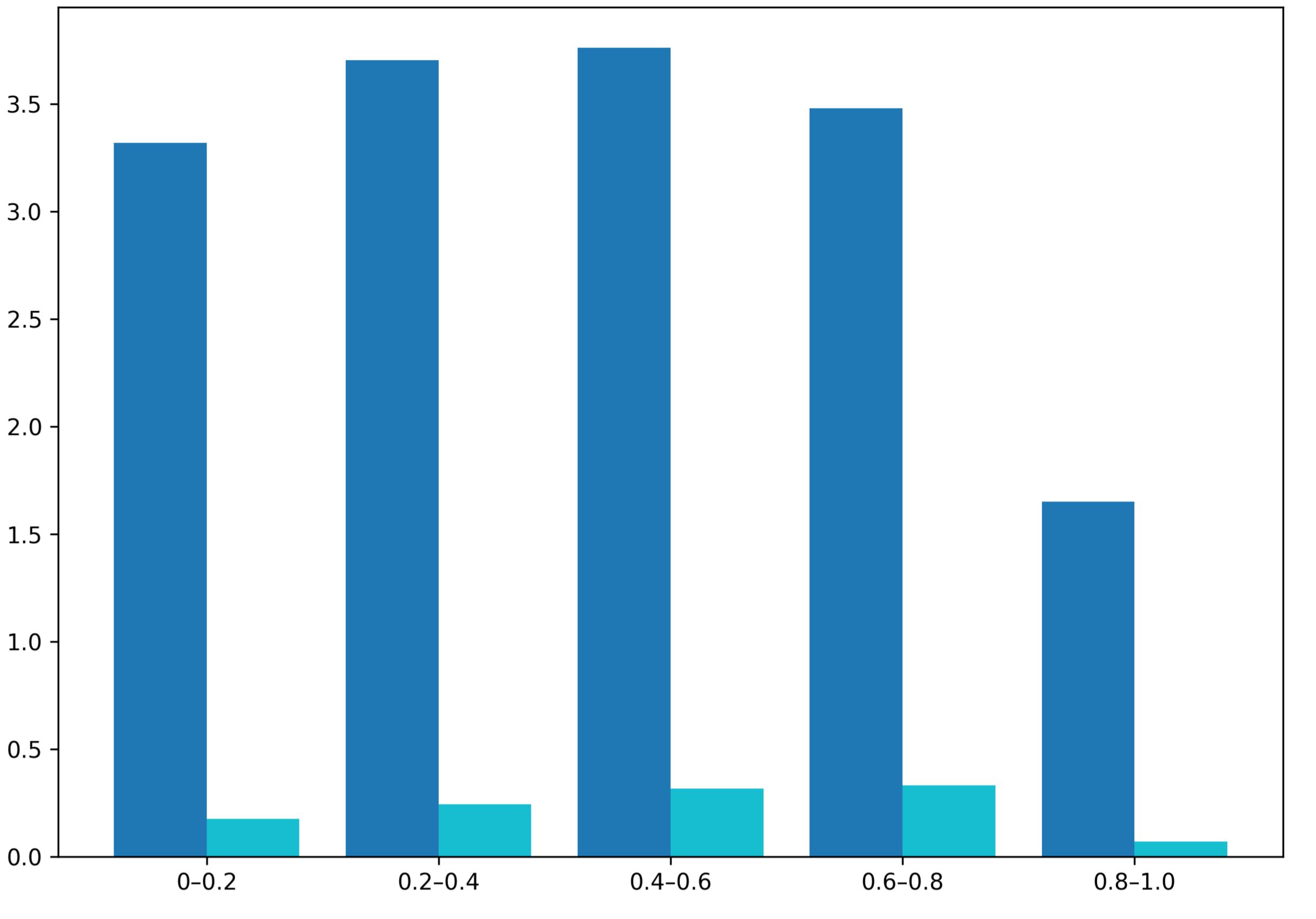}
    \caption{Suggestion}
\end{subfigure}
\hfill
\begin{subfigure}{0.45\textwidth}
    \centering
    \includegraphics[width=\linewidth]{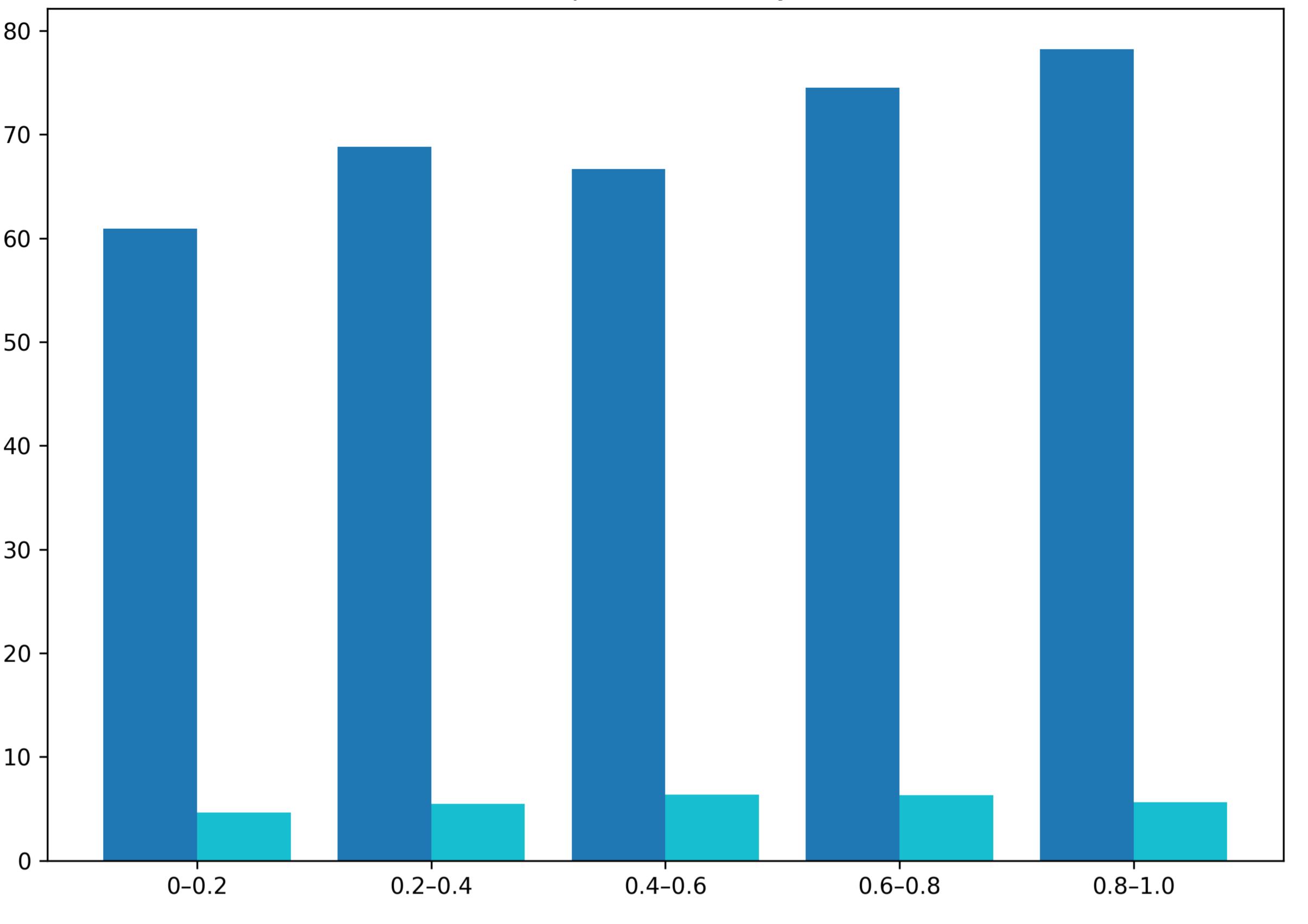}
    \caption{Probability}
\end{subfigure}

\caption{
Distribution of certainty and uncertainty linguistic aspects across overstatement score bins.
Each subfigure corresponds to a certainty category (Extent, Number, Framing, Condition, Suggestion, Probability).
Dark blue bars indicate certainty expressions, while light blue bars indicate uncertainty expressions.
}
\label{fig:uncertainty}
\end{figure*}

We investigate how varying degrees of overstatement influences the certainty of the claim \cite{pei-jurgens-2021-measuring}. Figure \ref{fig:uncertainty} shows that claims use mainly confident language with uncertainty aspects being uncommon, reflecting the standard that researchers present their findings with confidence and clarity \cite{salager1994hedges,cortes2004lexical}.

The observed patterns indicate that increased overstatement is driven by a greater use of probability certainty (e.g., \emph{will}, \emph{guaranteed}), extent certainty (e.g., \emph{fully}, \emph{completely}, \emph{exactly}), and number certainty (e.g., precise or absolute quantification), each of which serves to present claims as more definitive than is typically warranted. In contrast, framing certainty (e.g., \emph{show}, \emph{demonstrate}, \emph{verify}) exhibits a slight decline, suggesting a reduced reliance on evidential framing in favour of more assertive language. Suggestion and condition aspects occur in fewer than 5\% of claims and are therefore excluded from further analysis. These findings align with recent work on undergraduate theses, which reports a growing reliance on intensifiers across academic levels to strengthen claims; when paired with insufficient hedging, this tendency has been shown to undermine perceived credibility \cite{iftikhar2025hedging}. More broadly, this pattern is consistent with evidence that hedging language and other uncertainty-marking devices have become less common in academic writing in recent years \cite{wheeler2021more,yao2023promoting}.

\newpage
\onecolumn
\section{Prompts for data processing}\label{A:prompts}

\begin{table*}[h]
\centering
\small
\begin{tabular}{p{0.95\linewidth}}
\toprule
\texttt{You will be provided with the abstract and introduction of an academic paper along with a specific sentence from the paper. Your task is to determine whether the given sentence represents an original claim introduced by the authors that is directly relevant to the contribution or selling points of the paper.}\\
\\
\texttt{Labels:}\\
\texttt{original\_statement: The sentence explicitly presents a novel claim, finding, or result that is directly relevant to the key contributions of the paper. It reflects what the authors are aiming to promote or highlight as a significant contribution.}\\
\\
\texttt{not\_original\_statement: The sentence mainly provides background information, references prior work, describes common knowledge, or includes general context not directly tied to the unique contributions of the paper.}\\
\\
\texttt{The abstract and introduction of the paper:}\\
\texttt{Abstract:} \\
\texttt{\{ABSTRACT\}}\\
\\
\texttt{Introduction:} \\
\texttt{\{INTRODUCTION\}}\\
\\
\texttt{The sentence you are about to annotate:}\\
\texttt{\{SENTENCE\}}\\
\\
\texttt{You should:}\\
\texttt{1. Carefully review the context of the paper (abstract and introduction) and the given sentence. Then briefly justify whether the sentence is an original\_statement or not\_original\_statement (up to 100 words).}\\
\texttt{2. Provide the final annotation label in the format: <Label>\{your\_label\}<\//Label> }\\
\bottomrule
\end{tabular}
\caption{Prompt for Own statement labelling.}
\label{tab:prompt1}
\end{table*}

\begin{table*}[h]
\centering
\small
\begin{tabular}{p{0.95\linewidth}} 
\toprule
\texttt{You will be given a claim and a list of sentences. Your task is to identify the sentences that support the claim.}\\
\\
\texttt{A sentence supports the claim if it:}\\
\\
\texttt{- Directly provides evidence (e.g., experimental results, analysis, conclusions).}\\
\texttt{- Builds upon the claim by providing relevant context (e.g., background information).}\\
\\
\texttt{A supporting sentence must not:}\\
\\
\texttt{- Be a duplicate or paraphrase of the claim.}\\
\texttt{- Be incomplete.}\\
\texttt{- Contain text that appears to be part of an OCR-extracted table or figure (e.g., columns of numbers, symbols, "Table 1", or values not from a sentence). Such lines should always be ignored.}\\
\\
\texttt{The sentences are numbered, and you should return only the numbers of the supporting sentences.}\\
\\
\texttt{Claim:}\\
\texttt{\{CLAIM\}}\\
\\
\texttt{Sentences to evaluate:}\\
\texttt{\{NUMBERED SENTENCES\}}\\
\\
\texttt{Instructions:}\\
\texttt{Carefully review the claim and sentences. Provide a brief justification ($\leq$100 words) for which sentences support the claim.}\\
\texttt{If multiple sentences support the claim, list each number on a new line. If no sentences support the claim, return an empty <Label> tag.}\\
\texttt{Provide the final annotation label in the format:}\\
\texttt{<Label>}\\
\texttt{\{sentence numbers\}}\\
\texttt{</Label>}\\
\bottomrule
\end{tabular}
\caption{Prompt for text evidence extraction.}
\label{tab:prompt2}
\end{table*}

\begin{table*}[h]
\centering
\small
\begin{tabular}{p{0.95\linewidth}}
\toprule
\texttt{You will be provided with a research claim and a \{FIG\_TYPE\} (figure or table) extracted from an academic paper.}\\
\texttt{Your task is to determine whether the visual content is relevant to the claim --- that is, whether it provides evidence or context supporting the claim.}\\
\\
\texttt{A visual is relevant if it:}\\
\\
\texttt{- Directly provides evidence (e.g., experimental results, analysis, conclusions).}\\
\texttt{- Builds upon the claim by providing relevant context (e.g., background information).}\\
\\
\texttt{A visual is not relevant if it:}\\
\\
\texttt{- Contains no data or analysis tied to the claim.}\\
\texttt{- Shows unrelated or generic material.}\\
\texttt{- Is incomplete, unreadable, or too vague to judge its relevance.}\\
\\
\texttt{Labels:}\\
\texttt{relevant: The visual supports or builds upon the claim.}\\
\texttt{not\_relevant: The visual is unrelated to the claim.}\\
\\
\texttt{Claim:}\\
\texttt{\{CLAIM\}}\\
\\
\texttt{Visual information:}\\
\texttt{Type: \{FIG\_TYPE\}}\\
\texttt{Caption: \{CAPTION\}}\\
\texttt{Visible text: \{IMAGE\_TEXT\}}\\
\\
\texttt{Instructions:}\\
\texttt{1. Carefully review the claim and the visual.}\\
\texttt{2. Briefly justify ($\leq$100 words) whether the visual is relevant or not.}\\
\texttt{3. Provide the final label in this format:}\\
\texttt{<Label>\{relevant OR not\_relevant\}</Label>}\\
\bottomrule
\end{tabular}
\caption{Prompt for image-based evidence extraction, used for figures and tables.}
\label{tab:prompt3}
\end{table*}

\begin{table*}[h]
\centering
\small
\begin{tabular}{p{0.95\linewidth}} 
\toprule
\texttt{Your role is to assess the degree to which a claim is overstated based on the available evidence.} \\
\\
\texttt{``Overclaiming'' refers to rhetorical exaggeration: when the wording or framing of a claim amplifies its strength beyond what the paper’s own evidence supports.} \\
\texttt{It concerns rhetorical and linguistic inflation rather than factual correctness.} \\
\\
\texttt{The Input Information will include:} \\
\texttt{1. Original Claim: The claim under evaluation.} \\
\texttt{2. Evidence: Research findings, including figures, tables, or other relevant data supporting the claim.} \\
\\
\texttt{Optional. Review comment: Reviewer feedback relevant to the claim’s validity.} \\
\\
\texttt{Evaluate the claim against the provided evidence. Assign a score from 0 to 1 representing the degree of exaggeration using the following scale:} \\
\texttt{0.0: The claim contains no exaggeration and fully aligns with the evidence.} \\
\texttt{Values closer to 0: Minor exaggeration or slight over-interpretation.} \\
\texttt{Values closer to 1: Substantial exaggeration beyond what the evidence supports.} \\
\texttt{1.0: Major exaggeration or strong misrepresentation of the evidence.} \\
\\
\texttt{Justification: Provide a concise explanation that includes:} \\
\texttt{Instances of exaggerated wording, insufficient experiments, lack of experimental details, gaps in knowledge, weak grounding in evidence, or missing limitations.} \\
\texttt{Direct references to the relevant evidence supporting your reasoning.} \\
\texttt{If a review comment is included, consider relevant points but do not mention or reference the review.} \\
\texttt{Do not mention or restate the score in the justification.} \\
\\
\texttt{The claim to be assessed is:} \\
\texttt{\{CLAIM\}} \\
\\
\texttt{The review comment to be evaluated is:} \\
\texttt{\{REVIEW\}} \\
\\
\texttt{The evidence to be evaluated is:} \\
\texttt{\{EVIDENCE\}} \\
\\
\texttt{You should:} \\
\texttt{1. Review the claim and the text and image evidence. Summarize how the evidence influences your evaluation of the claim and briefly explain whether the claim is well-stated or overstated on the 0--1 scale (up to 100 words).} \\
\texttt{2. Provide the final score in the format: <score>\{score\}</score>.} \\
\texttt{3. Provide your justification in the format: <justification>\{justification\}</justification>.} \\
\bottomrule
\end{tabular}
\caption{Prompt for Overstatement label annotation.}
\label{tab:prompt4}
\end{table*}


\clearpage
\section{Case study evidence}\label{app:case_study}

\begin{table*}[!hp]
\centering
\small
\renewcommand{\arraystretch}{0.9}
\setlength{\tabcolsep}{2.9pt}

\begin{tabular*}{\textwidth}{@{\extracolsep{\fill}} p{\textwidth}}

\toprule
\textbf{Evidence (Input)} \\
\midrule
Geometrically, the embeddings of tokens are pushed toward the origin in the output layer of a model, compressing the dense regions in the semantic space toward the origin, making the embedding space more defined with concrete examples of words (see also Figure 1), instead of leaving many poorly-defined areas (Li et al., 2020). We provide a visualization of embedding geometry change in Figure 1. We suggest that this range plays a main role in making the entire semantic space isotropic. We find that temperature affects making embeddings isotropic: to push in-batch negatives to the lower bound, the temperature needs to be twice as large than to push them to the upper bound. Connecting this to our finding on high intra-sentence similarity, we observe that given a sentence/document-level input, certain semantic tokens drive the embeddings of all tokens to converge to a position, while functional tokens follow wherever they travel in the semantic space. Their performance on L2-norm is also well-aligned, again showing strong correlation between isotropy and L2-norm in the training process utilizing contrastive loss. Further, with limited space in the now compressed space, inputs have now learned to converge to one another to squeeze to a point while keeping its semantic relationship to other examples. We perform UMAP dimensionality reduction on embeddings provided by models up to 1000 step to preserve better global structure, and visualize only vanilla and 200-step embeddings. Specifically, for anisotropy baseline, temperature being too low even augments the vanilla model’s unideal behavior, and the same applies for L2-norm, by that temperature being too low actually pushes the embeddings even further from the origin. By contrast, mean pooling and max pooling demonstrate a faster convergence, with mean pooling being most promising on isotropy. For instance, removing the top 1 dominant dimension of minilmfinetuned seems to not affect the embeddings’ relative similarity to one another at all, preserving an $r^2$ of $.998$. Figure 1: Expanded semantic space produced by contrastive learning (CL), visualized with UMAP. At the beginning of training, all embeddings occupied a narrow cone. After 200 steps of fine-tuning with a contrastive loss, they spread out to define a larger semantic space. Firstly, we present the centered property we are measuring, anisotropy. We showed the theoretical promise of uniformity brought by contrastive learning through measuring anisotropy, complemented by showing the flattened domination of top dimensions. Higher self-similarity indicates less contextualization. The central question posed in this paper revolves around the mechanism involved in the contrastive learning process that diminishes anisotropy, leading to an isotropic model. Figure 7 further validates this through showing that higher temperatures compress the semantic space in general, pushing instances to the origin. Given a token x, we denote the set of token embeddings of x contextualized by different contexts in corpus S as SX. We find that with optimal hyperparameters, the representations go through less change after 200 steps. We first use the vanilla mpnet to encode the STSB subset we have constructed.
\\
\quad\quad\quad
\includegraphics[width=0.35\linewidth]{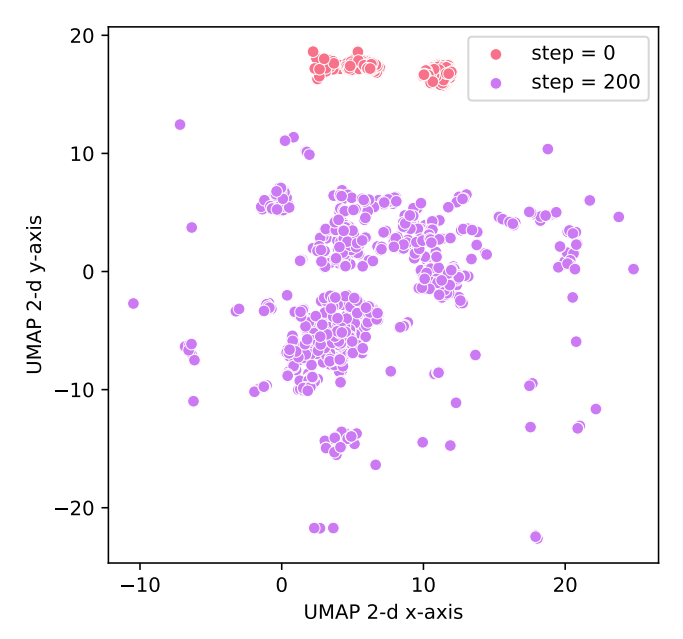}
\hfill
\includegraphics[width=0.5\linewidth]{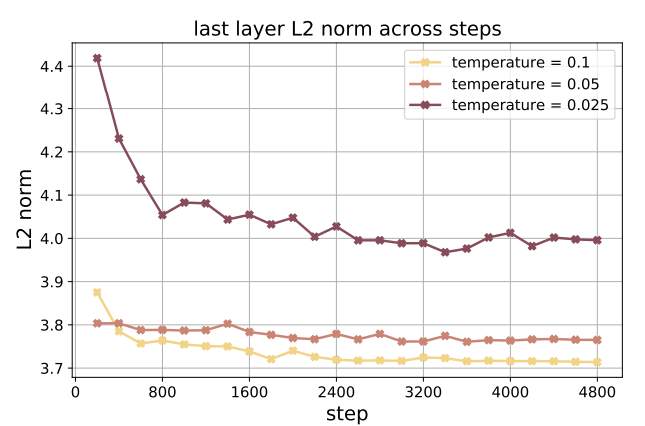}
\quad\quad\quad\\
\bottomrule
\end{tabular*}
\caption{Evidence utilised for case study retrieved using our fine-tuned Qwen reranker on the top 20 relevant sentences.}
\label{tab:casestudy_evidence}
\end{table*}

\end{document}